\documentclass[lettersize,journal]{IEEEtran}
\usepackage{amsmath,amsfonts}
\usepackage{algorithmic}
\usepackage{algorithm}
\usepackage{array}
\usepackage[caption=false,font=normalsize,labelfont=sf,textfont=sf]{subfig}
\usepackage{textcomp}
\usepackage{stfloats}
\usepackage{url}
\usepackage{verbatim}
\usepackage{graphicx}
\usepackage{cite}
\begin{document}

\title{SWIR-cGAN: Synthesizing SWIR Bands Using VNIR Imagery from CBERS and AMAZONIA-1}

\author{Alisson Cleiton de Oliveira, Rennan de Freitas Bezerra Marujo, Thales Sehn Körting

}



\maketitle

\begin{abstract}
Spectral band synthesis is an important research topic in remote sensing. The Brazilian satellites CBERS-4, CBERS-4A, and AMAZONIA-1 provide high temporal resolution (1--3 days) visible and near-infrared (VNIR) imagery through the Wide-Field Imager (WFI), but lack short-wave infrared (SWIR) bands. Recent deep learning methods have shown promising results for spectral synthesis, but challenges remain for cross-satellite generalization and spectrally distant bands. This paper presents a conditional Generative Adversarial Network (cGAN) with an attention-enhanced U-net generator and a multi-loss objective to synthesize Sentinel-2/MSI SWIR bands 11 (SWIR-1) and 12 (SWIR-2) from WFI VNIR inputs. Experiments used 4,125 patches for training and cross-validation and 1,026 unseen patches for testing, with hyperparameters optimized using a subset of the training data. On SWIR-1, the model achieved SSIM of $0.900$, PSNR of $31.44$~dB, and MAE of $0.019$ in TOA reflectance. On SWIR-2, it achieved SSIM of $0.917$, PSNR of $33.35$~dB, and MAE of $0.015$. Benchmarking was performed on the test dataset, with each model optimized separately prior to evaluation. SWIR-cGAN outperformed the baselines, particularly for SWIR-2. The ablation study identified the multi-loss as the main performance contributor, while the attention and residual modules provided additional improvements. Our results demonstrate the feasibility of synthesizing SWIR bands from WFI imagery. The SWIR-cGAN is available on https://github.com/marujore/swir-cgan.
\end{abstract}

\begin{IEEEkeywords}
Spectral band synthesis, Brazilian satellites, short-wave infrared, Wide-Field Imager (WFI), Sentinel-2.
\end{IEEEkeywords}

\section{Introduction}
\IEEEPARstart{T}{he} China-Brazil Earth Resources Satellite (CBERS) program ensures continuity of comparable data over time \cite{ref1}. Since 1999, this program has evolved to the current operational CBERS-4 and CBERS-4A satellites. Additionally, AMAZONIA-1, launched in 2021, is the first satellite fully developed and operated in Brazil. These platforms carry multiple optical sensors, including the Wide-Field Imager (WFI). The WFI sensors acquire visible (RGB) and near-infrared (NIR) imagery (hereafter VNIR) at revisit intervals of 1--3 days but lack a short-wave infrared (SWIR) band, which is important for environmental applications such as burned-area detection.

Approaches based on conditional Generative Adversarial Networks (cGANs) \cite{ref2} have been applied to reconstruct spectral bands from limited inputs. Rodríguez et al. \cite{ref3} evaluated U-net and ResNet generators with a PatchGAN \cite{ref4} discriminator to synthesize Sentinel-2/MSI bands from RGB channels. While their framework relied on visible wavelengths, incorporating NIR would be a practical choice, as it is readily available on most orbital sensors. Similarly, Illarionova et al. \cite{ref5} employed cGANs to predict NIR from WorldView-2 RGB images. They emphasized that NIR synthesis is constrained by the spectral information available in the visible bands. Green roofs and vegetation, for instance, may exhibit similar RGB profiles yet distinct NIR responses, highlighting the importance of spatial context, shape, and texture in spectral inference \cite{ref5}. Bridging more spectrally distant wavelengths, Hu et al. \cite{ref6} used GANs for synthetic aperture radar (SAR)-to-optical translation, aiming to support the mapping of burned areas. Although optical outputs avoid radar speckle, cross-domain translation remains difficult due to heterogeneous landscapes, class imbalance, and adversarial training instability. These studies demonstrate both the potential of GANs for spectral band synthesis and the inherent challenges of this generation.

Motivated by such works, we introduce SWIR-cGAN for synthesizing SWIR bands using VNIR bands from WFI. Our framework addresses spectral ambiguity, spatial degradation, and GAN instability through targeted design choices. Adding NIR to RGB bands provides richer information for predicting SWIR behavior. Attention gates and residual connections in the U-net generator help maintain fine textures and context, while the PatchGAN discriminator captures local details. A composite multi-loss formulation promotes training stability, preserves high-frequency details, and mitigates mode collapse when synthesizing SWIR bands (B11 and B12) from MSI.

The present method could support Brazil’s environmental monitoring programs, such as the Real-Time Deforestation Detection System (Deter), which issues daily deforestation and forest degradation alerts in the Amazon and Cerrado biomes through image analysis by experts. Deter recently transitioned from using 500 m MODIS images to 55 m (CBERS-4A) and 64 m (CBERS-4 and AMAZONIA-1) WFI imagery \cite{ref7}. 

\section{Dataset}

\begin{figure}[!ht]
\centering
\includegraphics[width=\columnwidth]{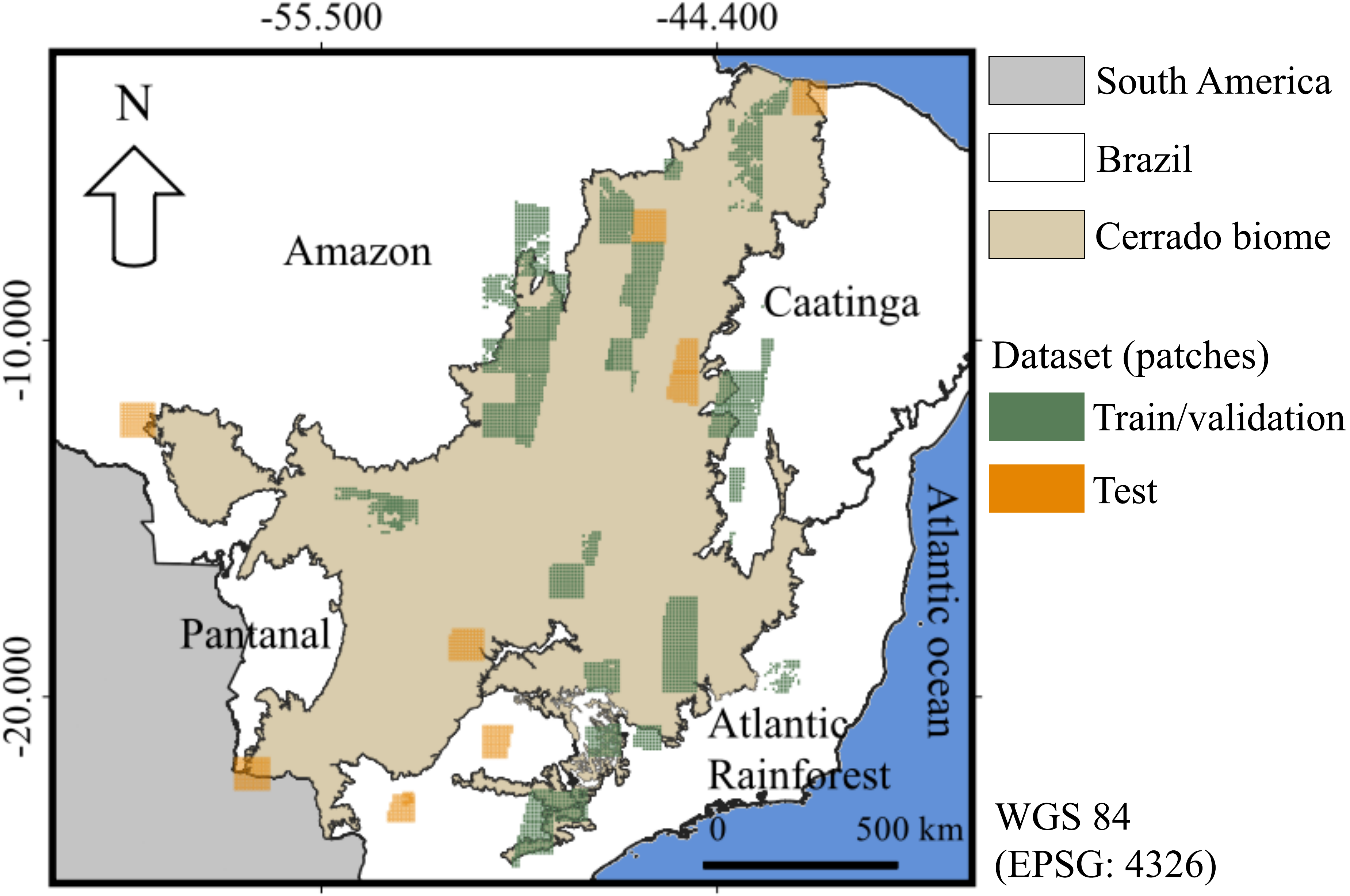}
\caption{Geographic distribution of the spatiotemporal patch pairs. The open dataset has 5,151 paired WFI--MSI patches acquired on the same day: 1,742 from CBERS-4A (1,361 train/val, 381 test), 1,796 from CBERS-4 (1,446 train/val, 350 test), and 1,613 from AMAZONIA-1 (1,318 train/val, 295 test).}
\label{fig_1}
\vspace{-5mm}
\end{figure}

The WFI images were converted to top-of-atmosphere (TOA) reflectance, while the Sentinel-2/MSI bands (Level-1C) were resampled to the WFI spatial resolution (55 m or 64 m) with bilinear interpolation. Using the \texttt{AROSICS} library \cite{ref8}, the MSI NIR band (B8) was coregistered with the WFI NIR reference, and its estimated displacement was applied to the SWIR bands. This improved spatial source–target correspondence, reducing misalignment effects and noises on band synthesis. Therefore, each spatiotemporal pair was acquired on the same day. The training and validation dataset comprises 4{,}125 pairs of $128 \times 128$-pixel patches from 10 distinct WFI scenes (Level-4), matching 65 distinct MSI tiles across nine dates in 2024. The test dataset comprises 1{,}026 patches with no spatial or temporal overlap with the training and validation subsets. Figure~\ref{fig_1} illustrates the spatial distribution of the patches, with most located in the Brazilian Cerrado biome.

\section{Methods}

\subsection{Network Architecture}

\begin{figure}[!ht]
\centering
\includegraphics[width=\columnwidth]{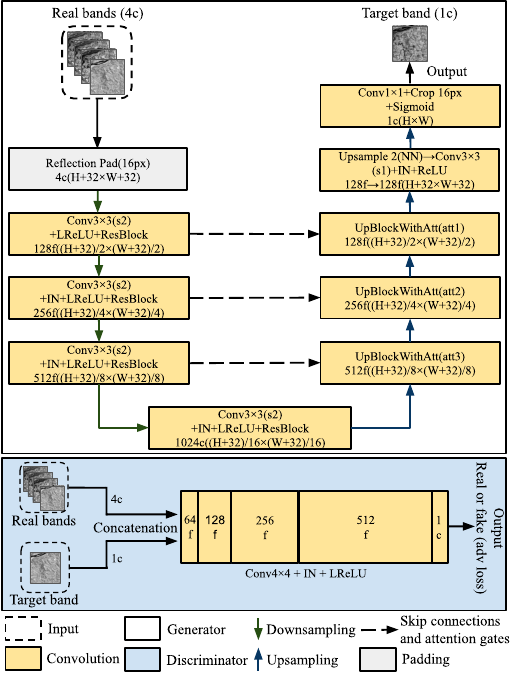}
\caption{Overview of the proposed attention U-net–based cGAN. 16px: added pixels; c: number of channels; H (height) and W (width): spatial dimensions; Conv: convolution; s2: stride = $2$; LReLU: Leaky ReLU; ResBlock: residual block; IN: instance normalization; f: number of filters; Att: attention; NN: nearest neighbor; s1: stride = $1$; adv loss: adversarial loss.} 
\label{fig_2}
\vspace{-4mm}
\end{figure}

Figure~\ref{fig_2} illustrates the SWIR-cGAN framework. The generator is based on a U-net architecture with attention gates and residual connections to improve feature propagation and focus on spatially relevant regions. It is trained with a multi-loss objective, where each loss component aims to address a distinct aspect of the synthesis problem. The implementation uses \texttt {PyTorch 2.7.1} \cite{ref9} with CUDA 11.8 support.

The $L_1$ loss enforces pixel-wise correspondence while reducing the oversmoothing often associated with $L_2$-based optimization. The Structural Similarity Index Measure (SSIM) loss promotes structural consistency by preserving local luminance, contrast, and structural patterns. The gradient loss preserves high-frequency spatial transitions, minimizing blurred boundaries between surface targets. The perceptual loss, computed with a pre-trained VGG-16 network, promotes feature-space similarity to improve textural patterns. Finally, the adversarial loss, guided by a PatchGAN discriminator, promotes distribution alignment with real SWIR bands, improving realism.

\subsubsection{Generator}

The generator follows an encoder--decoder U-net architecture enhanced with residual learning, attention gates, and reflection padding. Let $x \in \mathbb{R}^{4 \times H \times W}$ denote the input tensor (VNIR patches). To mitigate edge artifacts, symmetric reflection padding of size $p=16$ is applied:
\begin{equation}
x_{\text{pad}} = \text{ReflectPad}(x, p)
\end{equation}

\paragraph{Encoder}

The encoder consists of four downsampling stages. Each stage applies reflection padding ($p=1$) to minimize boundary artifacts, followed by a $3 \times 3$ convolution with stride $s=2$, instance normalization (IN) (except in the first stage) to accelerate convergence, and LeakyReLU (LReLU) activation ($\alpha=0.2$). This operation reduces the spatial resolution by a factor of two while increasing the number of feature channels (c) from $4$ to $1024$ in the bottleneck.

Each stage is followed by a residual block (ResBlock) composed of two convolutional layers ($3 \times 3$ convolution with stride $1$), each followed by instance normalization, with a ReLU activation after the first convolution and after the residual addition. The block output is added to the input via an identity skip connection:
\begin{equation}
\text{Res}(f) = \text{ReLU}\left(f + \text{IN}(W_2 * \text{ReLU}(\text{IN}(W_1 * f)))\right)
\end{equation}
where $*$ denotes convolution. This structure facilitates gradient flow and stabilizes training. The number of feature channels increases progressively as $4 \rightarrow 128 \rightarrow 256 \rightarrow 512 \rightarrow 1024$.

\paragraph{Decoder}  Each stage performs nearest-neighbor upsampling (scale factor $2$), with value preservation, followed by an attention gate that weights the corresponding encoder features to emphasize spatially relevant regions. The attended skip features are then concatenated with the upsampled features and processed by a convolutional block with reflection padding, instance normalization, and ReLU activation, followed by a residual block. The number of feature channels decreases from $1024$ to $128$. Dropout is then applied in the two deepest decoding layers to regularize the model and mitigate overfitting.

\paragraph{Attention Gates}

Skip connections are modulated by attention gates that suppress irrelevant encoder features. This mechanism enables the model to focus on spatially relevant regions. Each gate receives a gating signal $g$ from the decoder (coarse-scale features) and an encoder feature map $x$ (skip connection). The attention coefficients are computed as:
\begin{equation}
\alpha = \sigma \left( \psi \left( \text{ReLU}(W_g g + W_x x) \right) \right)
\end{equation}
where $W_g$ and $W_x$ are $1 \times 1$ convolutions projecting the inputs to 
an intermediate feature space, $\psi$ is a linear projection, and $\sigma$ 
denotes the sigmoid function. The attended feature map is obtained by:
\begin{equation}
x' = x \odot \alpha
\end{equation}
where $\odot$ denotes element-wise multiplication. This mechanism is applied at each of the three skip connections, enabling the network to selectively  propagate high-level contextual information while suppressing irrelevant background features.

\paragraph{Losses}
The total loss is a weighted sum of L1, SSIM, gradient, perceptual and adversarial components, with band-specific weights selected via hyperparameter search (Table~\ref{tab:hyperparams}).

\paragraph{Output Layer} 
The final layer (synthesized SWIR band) applies a $1 \times 1$ convolution to map $128$ feature channels to a single output band (target SWIR band), followed by a sigmoid activation. The initial reflection padding ($p=16$) is removed from the output, restoring the original spatial resolution and ensuring consistency with the input patch.

\subsubsection{Discriminator}
The discriminator is a fully convolutional PatchGAN that operates on the concatenated image pair $[x, y] \in \mathbb{R}^{5 \times H \times W}$, combining the 4-band input $x$ and the target band $y$. It is trained with a binary cross-entropy (BCE) loss to distinguish real pairs from generated ones. Label smoothing is applied, using targets of $0.9$ for real samples and 0 for synthesized samples to stabilize adversarial training. The network consists of a sequence of convolutional layers with kernel size $4 \times 4$, using stride $s=2$ in the initial layers and $s=1$ in the deeper layer, followed by instance normalization (except in the first layer) and LeakyReLU activation ($\alpha=0.2$). The number of feature channels increases progressively from $64$ to $128$, $256$, and $512$. The network outputs a patch-wise map, where each element indicates the probability of a real patch, guiding the generator toward realistic SWIR synthesis.

\subsection{Experiment}

All processing was conducted in the Brazil Data Cube Lab (BDC-Lab), a geospatial data science environment hosted on the infrastructure of Brazil's National Institute for Space Research (INPE) that enables virtual machine (VM) instantiation with customizable configurations \cite{ref10}. For this experiment, a VM was configured with two NVIDIA Tesla V100 GPUs (16 GB each) and up to 232 GB of system RAM.

A random $20\%$ subset of the training dataset was used for hyperparameter tuning using \texttt{Optuna} \cite{ref11}. Hyperparameters were optimized independently for each SWIR band to account for its spectral characteristics, using $50$ trials of $25$ epochs each and a batch size of $16$. Our cGAN uses a five-component loss objective with individually tuned weights, optimized learning rates (generator and discriminator), and dropout.

Using the optimized hyperparameters, five-fold cross-validation was performed on the remaining training data with a fixed random seed of $42$. Each fold used disjoint training and validation partitions, ensuring every sample was evaluated exactly once. On-the-fly data augmentation was performed using \texttt{Albumentations}\cite{ref12}, including horizontal and vertical flips and $90^\circ$ rotations, each applied with a probability of $0.5$.

To benchmark the proposed method, we compare it against three architectures representing distinct design paradigms: U-net \cite{ref13}, a purely convolutional baseline; ESRGAN \cite{ref14}, a state-of-the-art super-resolution architecture adapted for spectral synthesis with $12$ RRDB blocks, $32$ feature channels, and $16$ growth channels, incorporating relativistic adversarial learning; and ViT-U-net, which replaces the convolutional bottleneck with a Vision Transformer (ViT) to capture long-range dependencies \cite{ref15},\cite{ref16}, with a PatchGAN discriminator.

All models underwent the \texttt{Optuna} hyperparameter optimization, with search spaces tailored to each architecture. For U-net, we optimized learning rate and dropout. For ESRGAN, we optimized L1, perceptual, and GAN loss weights, learning rates, dropout, and warmup schedule. For ViT-U-net, we optimized L1 and GAN losses, learning rates, and dropout.

Each model was trained on $95\%$ of the data, reserving $5\%$ for early stopping and learning-rate scheduling. Training was performed under identical conditions, including fixed random seeds, deterministic CUDA operations, and data splits, ensuring that performance differences reflect architectural choices rather than training variability. Generalization was evaluated on a test dataset comprising 1,026 unseen WFI--MSI patches.

The ablation study then assessed the contributions of architectural components and loss functions in SWIR-cGAN under the same protocol. Results were evaluated using SSIM, Peak Signal-to-Noise Ratio (PSNR), Pearson correlation, Mean Absolute Error (MAE), and the Wilcoxon signed-rank test. 

\section{Results and discussion}

\subsection{Hyperparameter optimization}

The optimized hyperparameters in Table~\ref{tab:hyperparams} reveal distinct training strategies for the SWIR-1 and SWIR-2 bands, reflecting differences in spectral complexity and synthesis difficulty.

\begin{table}[!ht]
\caption{Optimized hyperparameters for the SWIR-cGAN model.}
\label{tab:hyperparams}
\centering
\begin{tabular}{lccc}
\hline
\textbf{Parameter} & \textbf{Search space} & \textbf{SWIR-1} & \textbf{SWIR-2} \\
\hline
Adversarial loss            & $[0.01, 1.0]$          & 0.019  & 0.025  \\
L1 loss                     & $[1.0, 20.0]$          & 13.627 & 19.556 \\
SSIM loss                   & $[1.0, 50.0]$          & 42.781 & 34.991 \\
Gradient loss               & $[10.0, 200.0]$        & 133.933 & 180.192 \\
Perceptual loss             & $[0.0, 20.0]$          & 1.213  & 6.532 \\
Generator lr     & $[10^{-5}, 5\times10^{-4}]$ & $4.87\times10^{-4}$ & $4.99\times10^{-4}$ \\
Discriminator lr & $[10^{-5}, 5\times10^{-4}]$ & $1.10\times10^{-5}$ & $1.00\times10^{-5}$ \\
Dropout                     & $[0.0, 0.2]$           & 0.065  & 0.129  \\
\hline
{\footnotesize lr: learning rate.}
\end{tabular}
\end{table}

SWIR-1 is driven primarily by structural objectives, with SSIM and gradient losses dominating over pixel-wise and adversarial terms, indicating a stronger emphasis on spatial consistency. SWIR-2 adopts a more balanced multi-objective setup, with increased contributions from L1, perceptual, adversarial, and gradient losses, together with stronger regularization, suggesting a more complex and less directly recoverable mapping. Both cases use a higher generator learning rate than the discriminator, supporting stable adversarial optimization.

\subsection{Cross-validation}

Table~\ref{tab:results} summarizes the validation performance for the synthesized SWIR bands over the five cross-validation folds.

\begin{table}[!ht]
\centering
\caption{Cross-validation performance for SWIR synthesis. Values are reported as mean $\pm$ standard deviation over folds.}
\label{tab:results}
\begin{tabular}{cccc}
\hline
 & \textbf{SSIM} & \textbf{PSNR} & \textbf{MAE} \\ \hline
SWIR-1 & $0.900\pm0.002$ & $31.44\pm0.1$ & $0.019\pm0.0002$ \\
SWIR-2 & $0.917\pm0.002$ & $33.35\pm0.1$ & $0.015\pm0.0002$ \\ \hline
\end{tabular}
\end{table}

SWIR-1 centered at 1,610 nm achieved an SSIM of $0.900$, PSNR of $31.44$ dB, and MAE of $0.019$, indicating moderate synthesis quality, corresponding to average absolute errors of approximately $1.9\%$ in TOA reflectance (a dimensionless quantity). The best models were obtained after approximately 81 training epochs, with generator and discriminator losses decreasing from $11.5$ to $7.8$ and from $0.67$ to $0.33$, respectively.

SWIR-2 achieved superior performance across all metrics (SSIM: $0.917$; PSNR: $33.35$~dB; MAE: $0.015$) after convergence at around 80 epochs, with generator loss decreasing from $12.1$ to $8.64$ and discriminator loss from $0.67$ to $0.38$. Despite outperforming SWIR-1 in synthesis metrics, the higher generator loss suggests greater difficulty in modeling fine spectral details. Low cross-fold variance indicates systematic rather than data-driven differences.

These results show a decoupling between adversarial and synthesis metrics: SWIR-1 has lower generator loss but reduced synthesis accuracy, while SWIR-2 shows the opposite pattern. This may reflects a trade-off between realism and structural fidelity that varies across both spectral bands. Despite these dynamics, stable cross-fold performance validates the SWIR-cGAN for Sentinel-2/MSI SWIR synthesis from WFI VNIR inputs.

\subsection{Baselines}

Table~\ref{tab:benchmark} compares SWIR-cGAN with other methods. Across both bands, SWIR-cGAN achieves the best performance, with the highest SSIM, PSNR, and Pearson correlation, and the lowest MAE. The performance gap over baselines is more pronounced for SWIR-2, suggesting better handling of its higher spectral complexity. Among baselines, ESRGAN performs fairly but falls short in correlation and error metrics (Pearson: $0.83$ for ESRGAN vs. $0.87$ for SWIR-cGAN).

\begin{table*}[!ht]
\centering
\caption{Benchmark comparison. Values are reported as mean$\pm$standard deviation over the test set.}
\label{tab:benchmark}
\renewcommand{\arraystretch}{1.2}
\setlength{\tabcolsep}{6pt}

\begin{tabular}{lcccccccc}
\hline
\textbf{Architecture} &
\multicolumn{4}{c}{\textbf{SWIR-1}} &
\multicolumn{4}{c}{\textbf{SWIR-2}} \\
\cline{2-9}
&
\textbf{SSIM} &
\textbf{PSNR} &
\textbf{Pearson} &
\textbf{MAE} &
\textbf{SSIM} &
\textbf{PSNR} &
\textbf{Pearson} &
\textbf{MAE} \\
\hline

SWIR-cGAN &
\textbf{0.87$\pm$0.09} &
\textbf{29.28$\pm$3.7} &
\textbf{0.87$\pm$0.20} &
\textbf{0.03$\pm$0.02} &
\textbf{0.86$\pm$0.10} &
\textbf{30.15$\pm$4.2} &
\textbf{0.87$\pm$0.17} &
\textbf{0.03$\pm$0.02} \\

ESRGAN &
0.85$\pm$0.09 &
28.12$\pm$3.9 &
0.83$\pm$0.20 &
0.03$\pm$0.02 &
0.84$\pm$0.10 &
28.04$\pm$4.1 &
0.83$\pm$0.21 &
0.04$\pm$0.02 \\

ViT-U-net &
0.82$\pm$0.13 &
26.61$\pm$4.7 &
0.81$\pm$0.20 &
0.05$\pm$0.04 &
0.81$\pm$0.14 &
27.08$\pm$4.5 &
0.83$\pm$0.22 &
0.04$\pm$0.02 \\

U-net &
0.81$\pm$0.14 &
26.17$\pm$5.0 &
0.81$\pm$0.20 &
0.05$\pm$0.04 &
0.82$\pm$0.13 &
27.56$\pm$4.3 &
0.82$\pm$0.21 &
0.04$\pm$0.02 \\

\hline
\end{tabular}
\end{table*}

U-net and ViT-U-net yield lower SSIM ($0.81$ and $0.82$, respectively) and higher MAE ($0.04$ and $0.05$). ViT-U-net does not reliably outperform U-net; for SWIR-2, U-net even surpasses it (SSIM: $0.82$ vs. $0.81$). This suggests that convolutional inductive biases are more suitable than global attention for this task, as transformers may dilute local spectral correlations. ViT-U-net might also demand more training data.

All methods present high standard deviations for SWIR-1 (SSIM: $0.09$ to $0.14$; PSNR: $3.7$ to $5.0$~dB) and SWIR-2 (SSIM: $0.10$ to $0.14$; PSNR: $4.1$ to $4.5$~dB), reflecting some synthesis difficulty and variability, possibly related to differences between CBERS-4, CBERS-4A, and AMAZONIA-1 WFI radiometry and land-cover types. SWIR-cGAN, however, offers the best trade-off between performance and stability.

\begin{figure}[!ht]
\centering
\includegraphics[width=\columnwidth]{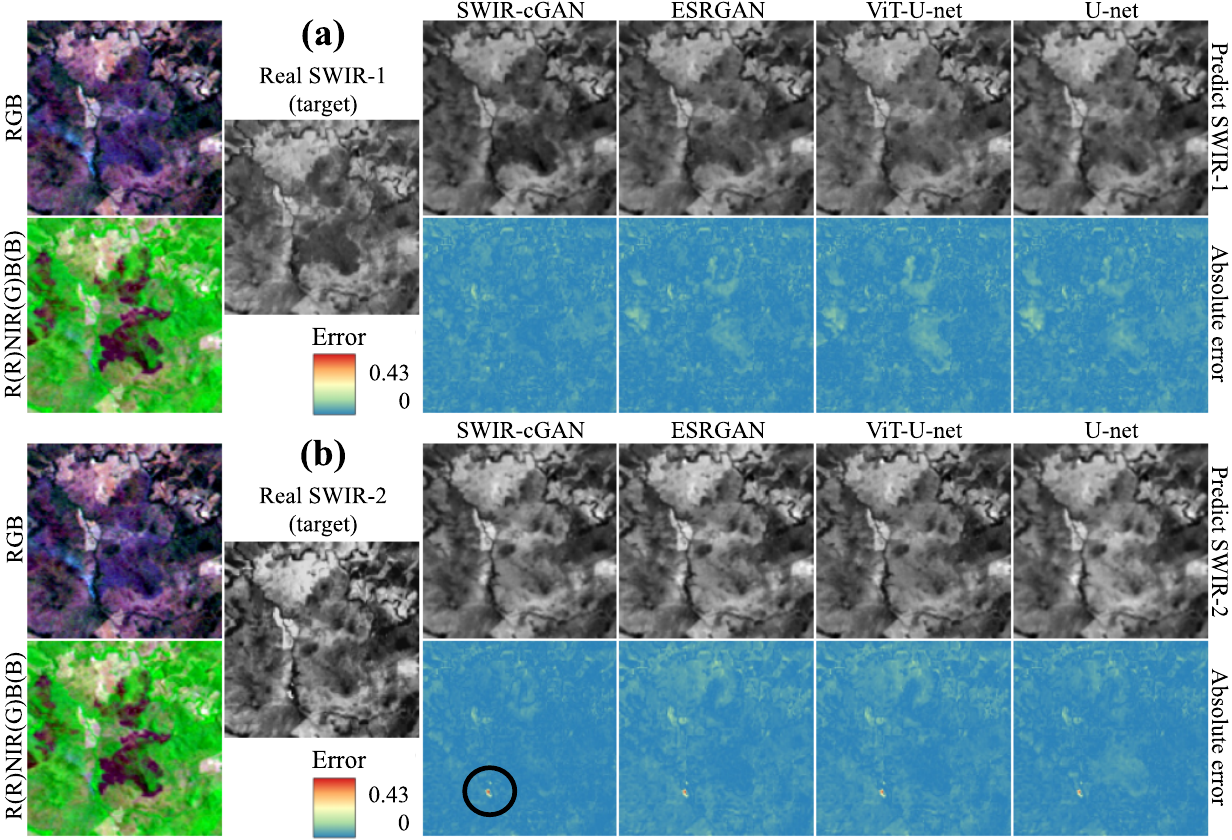}
\caption{Examples of image patches containing vegetation and burned areas (represented by dark tones at the center of the false-color composites). (a) Predicted SWIR-1 band; (b) Predicted SWIR-2 band. There is an active fire (black circle) in the southern portion with higher errors in SWIR-2 band.} 
\label{fig_3}
\vspace{-4mm}
\end{figure}

Figure~\ref{fig_3} shows synthesized SWIR bands for a patch with vegetation and fire-affected regions. Figure~\ref{fig_3}a highlights lower errors for SWIR-cGAN, mainly over burned areas. Figure~\ref{fig_3}b shows a small region with high errors related to an active fire.

\subsection{Ablation study}

Table~\ref{tab:ablation} presents the contribution of each component to SWIR synthesis across target bands. The full model achieves the best performance for both SWIR-1 and SWIR-2. For SWIR-1, removing any component significantly degrades SSIM and PSNR, with most effects being highly significant ($p < 0.001$). For SWIR-2, the multi-loss and attention mechanism show the strongest impact ($p < 0.001$ for SSIM), while reflection padding and residual blocks have less influence.

\begin{table*}[!ht]
\centering
\caption{Ablation study results. Values are reported as mean $\pm$ standard deviation. Significance levels indicate statistical differences with respect to the Full model (one-sided Wilcoxon signed-rank test).}
\label{tab:ablation}
\renewcommand{\arraystretch}{1.2}
\setlength{\tabcolsep}{9pt}

\begin{tabular}{lcccccccc}
\hline
\textbf{Configuration} &
\textbf{Attention} &
\textbf{Residual} &
\textbf{\begin{tabular}[c]{@{}c@{}}Reflection\\padding\end{tabular}} &
\textbf{\begin{tabular}[c]{@{}c@{}}Multi-\\loss\end{tabular}} &
\multicolumn{2}{c}{\textbf{SWIR-1}} &
\multicolumn{2}{c}{\textbf{SWIR-2}} \\
\cline{6-9}
&
&
&
&
&
\textbf{SSIM} &
\textbf{PSNR} &
\textbf{SSIM} &
\textbf{PSNR} \\
\hline

Full model &
$\checkmark$ &
$\checkmark$ &
$\checkmark$ &
$\checkmark$ &
\textbf{0.868$\pm$0.09} &
\textbf{29.28$\pm$3.7} &
\textbf{0.864$\pm$0.10} &
\textbf{30.20$\pm$4.2} \\

w/o Attention &
$\times$ &
$\checkmark$ &
$\checkmark$ &
$\checkmark$ &
0.866$\pm$0.09\rlap{$^{***}$} &
29.25$\pm$3.7\rlap{$^{*}$} &
0.861$\pm$0.10\rlap{$^{***}$} &
30.00$\pm$4.4\rlap{$^{***}$} \\

w/o Residual &
$\checkmark$ &
$\times$ &
$\checkmark$ &
$\checkmark$ &
0.864$\pm$0.09\rlap{$^{***}$} &
28.92$\pm$4.2\rlap{$^{***}$} &
0.863$\pm$0.09\rlap{$^{***}$} &
30.10$\pm$3.9\rlap{$^{*}$} \\

w/o Reflection padding &
$\checkmark$ &
$\checkmark$ &
$\times$ &
$\checkmark$ &
0.864$\pm$0.09\rlap{$^{***}$} &
28.84$\pm$3.9\rlap{$^{***}$} &
0.862$\pm$0.10\rlap{$^{***}$} &
30.20$\pm$4.0\rlap{$^{\text{ns}}$} \\

w/o Multi-loss &
$\checkmark$ &
$\checkmark$ &
$\checkmark$ &
$\times$ &
0.852$\pm$0.10\rlap{$^{***}$} &
27.93$\pm$5.1\rlap{$^{***}$} &
0.854$\pm$0.11\rlap{$^{***}$} &
29.66$\pm$4.5\rlap{$^{***}$} \\

\hline
\multicolumn{9}{l}{\footnotesize w/o: without; Significance levels: $^{\text{ns}}$ not significant ($p\text{--value} \ge 0.05$), $^{*}$ $p\text{--value} < 0.05$, $^{***}$ $p\text{--value} < 0.001$.}
\end{tabular}
\vspace{-3mm}
\end{table*}

Among the evaluated components, the multi-loss has the largest contribution to synthesis quality, as its removal decreases SSIM by $0.016$ ($1.8\%$) and PSNR by $1.35$ dB ($4.6\%$) for SWIR-1, with smaller but systematic reductions observed for SWIR-2 (SSIM: $-0.010$; PSNR: $-0.540$ dB). At $\alpha < 0.001$, the multi-loss is the only component showing significant effects on SSIM and PSNR in both SWIR bands.

Attention and residual mechanisms offer more modest improvements. Removing attention causes slight performance degradation, indicating its role in feature refinement rather than core synthesis. Residual connections have a larger impact on SWIR-1 (SSIM: $-0.004$, PSNR: $-0.360$ dB) than SWIR-2 (SSIM: $-0.001$; PSNR: $-0.100$ dB). Reflection padding has the least impact on quality among the evaluated components.

\section{Conclusion}
This letter demonstrates the feasibility of synthesizing SWIR bands from VNIR data using SWIR-cGAN. The results show that synthesis performance is band-dependent, with different spectral targets benefiting from distinct trade-offs among structural, perceptual, and adversarial objectives. Ablation experiments indicate that the multi-loss formulation has the greatest impact on synthesis quality among the evaluated design choices, while attention mechanisms, residual connections, and reflection padding provide additional improvements. The main limitation is the size and diversity of the training dataset, which may restrict generalization across geographic regions, seasons, and land-cover types. Although SWIR-cGAN may be applicable to data acquired at different processing levels and spatial resolutions, the experiments reported here and the proposed configuration specifically address WFI–MSI data under the acquisition and processing conditions adopted in this study. Finally, SWIR synthesis should be viewed as a means of augmenting VNIR observations for downstream environmental applications, including burned-area mapping and land-cover change detection, rather than as an end in itself.

This framework is relevant to operational monitoring programs such as Deter. Because WFI offers wide-swath coverage (approximately $800 \times 800$ km) and a high revisit frequency, synthesizing SWIR bands could, for instance, support the detection of degradation caused by fires. In the context of Deter, tests with human analysts could determine whether synthetic SWIR bands improve visual target interpretation alongside automated detection. Future efforts will focus on training SWIR-cGAN on an expanded dataset of WFI--MSI pairs across broader geographic and seasonal conditions to improve generalization to diverse land use and cover classes.

\section*{Data availability}
The open dataset is available on Science Data Bank (ScienceDB) at: https://doi.org/10.57760/sciencedb.34763. The SWIR-cGAN implementation and pre-trained weights are available at https://github.com/marujore/swir-cgan.

\section*{Acknowledgment}
The authors thank the Brazilian National Council for Scientific and Technological Development (CNPq, grants n. 302205/2023-3 and n. 142319/2025-2).


\end{document}